\def\year{2021}\relax
\documentclass[letterpaper]{article} 
\usepackage{aaai20}  
\usepackage{times}  
\usepackage{helvet} 
\usepackage{courier}  
\usepackage[hyphens]{url}  
\usepackage{graphicx} 
\usepackage{subfigure}
\usepackage{amsmath}
\usepackage{multirow}
\usepackage{rotating}
\usepackage{algpseudocode}
\usepackage{algorithm}

\newtheorem{definition}{Definition}
\newtheorem{problemdefinition}{Problem Definition}

\floatname{algorithm}{Algorithm}

\usepackage{graphicx}  
\title{Deep Multi-View Channel-Wise Spatio-Temporal Network \\for Traffic Flow Prediction}
\author{Hao Miao\textsuperscript{\rm 1}, Senzhang Wang\textsuperscript{\rm 1}, Meiyue Zhang\textsuperscript{\rm 1}, Diansheng Guo\textsuperscript{\rm 2}, Funing Sun\textsuperscript{\rm 2}, Fan Yang\textsuperscript{\rm 2}\\ 
\textsuperscript{\rm 1}Nanjing University of Aeronautics and Astronautics, \textsuperscript{\rm 2}Tencent\\ 
\textsuperscript{\rm 1}\{miaohao,szwang,meiyuezhang\}@nuaa.edu.cn, \textsuperscript{\rm 2}\{dguo,funingsun,wendellyang\}@tencent.com 
}
\begin{document}

\maketitle

\begin{abstract}
Accurately forecasting traffic flows is critically important to many real applications including public safety and intelligent transportation systems. The challenges of this problem include both the dynamic mobility patterns of the people and the complex spatial-temporal correlations of the urban traffic data. Meanwhile, most existing models ignore the diverse impacts of the various traffic observations (e.g. vehicle speed and road occupancy) on the traffic flow prediction, and different traffic observations can be considered as different channels of input features. We argue that the analysis in multiple-channel traffic observations might help to better address this problem. In this paper, we study the novel problem of multi-channel traffic flow prediction, and propose a deep \underline{M}ulti-\underline{V}iew \underline{C}hannel-wise \underline{S}patio-\underline{T}emporal \underline{Net}work (MVC-STNet) model to effectively address it. Specifically, we first construct the localized and globalized spatial graph where the multi-view fusion module is used to effectively extract the local and global spatial dependencies. Then LSTM is used to learn the temporal correlations. To effectively model the different impacts of various traffic observations on traffic flow prediction, a channel-wise graph convolutional network is also designed. Extensive experiments are conducted over the PEMS04 and PEMS08 datasets. The results demonstrate that the proposed MVC-STNet outperforms state-of-the-art methods by a large margin.
\end{abstract}
\section{Introduction}
As a typical spatio-temporal prediction task, traffic prediction has drawn considerable research attention in recent years due to the increasing amount of urban traffic and its significant impacts on real-world application. Accurate  traffic  forecasting  is  particularly  useful  to support Intelligent Transportation Systems (ITS), and can facilitate many real applications such as travel route planning\cite{huang2020multi}, and travel time estimation\cite{wang2018will}. 

Traditional statistics-based prediction approaches, such as ARIMA \cite{williams2003modeling} and VAR \cite{chandra2009predictions} have been widely used in road segment-level traffic flow prediction, and achieved promising performance. Predicting the traffic flows over a large road network, however, is a much more difficult task due to the very complex and non-linear spatio-temporal dependencies among the traffic data over different road links. Thus statistics-based models become less effective to handle the road network-level traffic prediction. With the recent advances of deep learning techniques, various deep learning models \cite{wang2020deep} are employed for traffic flow prediction and has achieved remarkable performance gains. ST-ResNet \cite{zhang2017deep} was proposed to collectively forecast the inflow and outflow in each region of a city. \cite{yao2019revisiting} proposed a Spatio-Temporal Dynamic Network (STDN) model for road network based traffic prediction. Diffusion convolution recursive neural network (DCRNN) \cite{li2018diffusion} integrated diffusion convolution and Seq2Seq structure for traffic flow prediction. Graph WaveNet \cite{wu2019graph} combined GCN with dlated casual convolution to capture the spatial-temporal dependencies of the traffic flows. STGCN \cite{yu2018spatio} used ChebNet graph convolution and 1D convolution to forecast the traffic flows in each road of a road network. GMAN \cite{zheng2020gman} proposed to utilize the attention mechanism to effectively extract spatial and temporal features for traffic prediction. 

However, existing works mainly focus on capturing the local spatial correlations of the traffic data, which follows the First Law of Geography \cite{tobler1970computer}: "\textit{Near things are more related than distant things}", but cannot fully reflect the global spatial dependencies. For example, previous work \cite{yao2018deep} showed that two regions with similar POI distribution or functionality(i.e., both commercial area), even though they are not geographically close to each other, can present very similar patterns on the semantic space of the spatio-temporal data (e.g., traffic flow). Such global spatial dependencies are not carefully considered by existing deep learning based approaches.
\begin{figure} \centering    
\subfigure[Traffic low $vs$ Vehicle speed] {
 \label{speed}     
\includegraphics[scale=0.25]{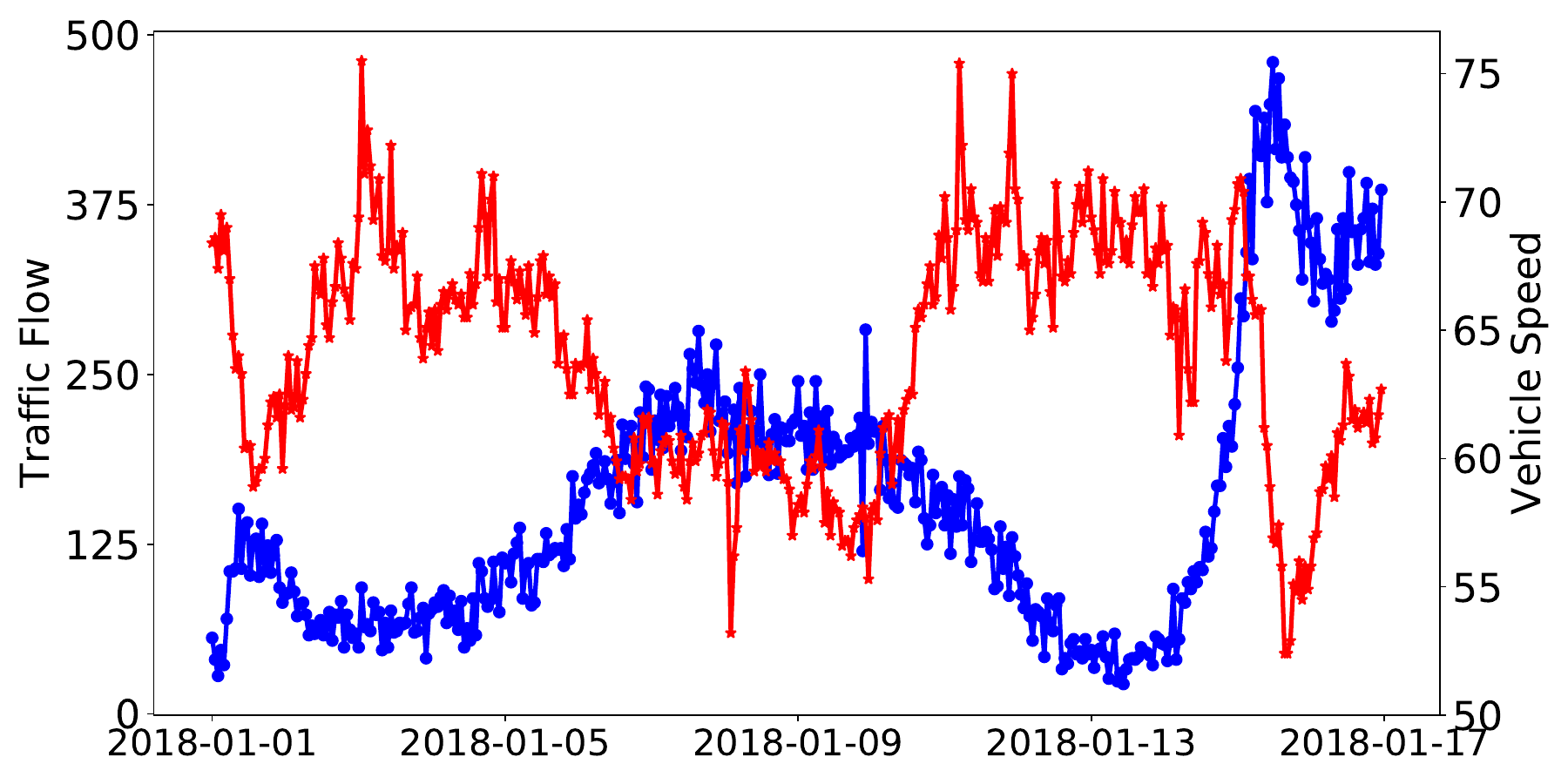} 
}     
\subfigure[Traffic flow $vs$ Road occupancy] { 
\label{occupy}     
\includegraphics[scale=0.25]{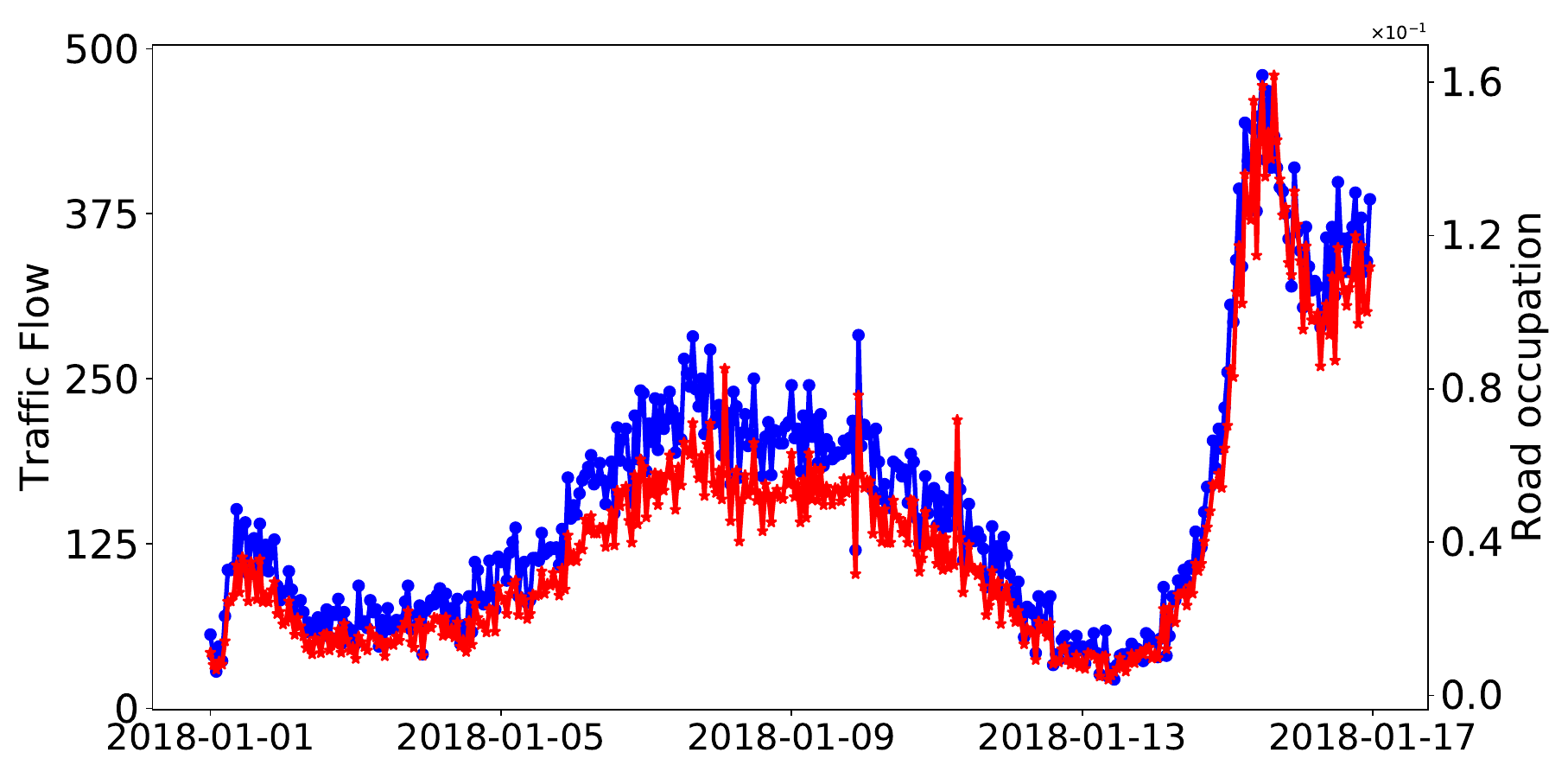}     
}    
\caption{Relationship between traffic flow and two traffic observation features (vehicle speed and road occupancy)}     
\label{relation} 
\end{figure}
Another limitation of existing methods is that the diverse impacts of the traffic observations on the traffic flow prediction task is largely ignored. As shown in Figure \ref{relation}, different types of traffic observations may have different impacts on traffic flows. Figure \ref{speed} shows the relationship between traffic flow and traffic speed. From the picture\ref{speed}, one can see that traffic flow and speed are negatively correlated. That is to say the traffic speed will slow down with the increase of traffic flows, and vice versa. Figure \ref{occupy} shows the relationship between traffic flow and the road occupancy where the change of road occupancy follows the change of traffic flows. Both traffic speed and road occupancy have impacts on traffic flows, but the two relationships are quite different which is like the idea in MMRate\cite{wang2014mmrate}. However, it is not considered in previous works how to model the influence of different traffic observation on the task of traffic flow prediction.

To address the above issues, in this paper we propose a deep multi-view channel-wise spatial-temporal network model named MVC-STNet for traffic flow prediction. To model the different relations between various input channels representing traffic flow, road occupy and vehicle speed, and the prediction, channel-wise graph convolutional network (CGCN) is proposed. CGCN can learn the data representations of each channel first, and then fuse them in a parametric-matrix-based way \cite{zhang2017deep}. To effectively capture the local and global spatial correlations, localized and globalized spatio-temporal graphs are constructed to learn the two spatial representations, and then a multi-view fusion module is proposed to fuse the localized and globalized data representations. Additionally, LSTM is also used to learn the sequential dependency of the traffic flows. Considering the external context features including holidays and weather can also significantly influence traffic flows, the external features are also incorporated into our model. We summarize our main contributions as follows:
\begin{itemize}
    \item A novel deep learning framework MVC-STNet is presented to perform spatio-temporal knowledge extraction for traffic flow prediction. By considering the spatial and temporal features exhaustively, the proposed model can effectively capture the complex spatial and temporal correlations.
    \item We propose a channel-wise graph neural network that can adaptively learn the different influences of input channels on traffic prediction. To the best of our knowledge, this is the first graph neural network based model for traffic flow prediction that considers the divergence between different input channels.
    \item We conduct experiments on two real world traffic datasets. The results show that our model is consistently and significantly better than existing state-of-the-art methods.
\end{itemize}
The remainder of the paper is organized as follows. Section 2 will review related work. Section 3 will give a formal problem definition.  Section 4 will first show the model framework and then introduce the model in detail. Evaluations are given in Section 5, followed by the conclusion in Section 6.

\section{Related Work}
This work is highly relevant to the research topics of traffic flow prediction and graph convolutional network. Next, we will review related works from the above two aspects.

\textbf{Traffic flow prediction.} As a typical spatio-temporal prediction task, traffic flow prediction has been studied for decades in intelligent transportation systems \cite{wang2020deep,yao2019revisiting}. Traditionally, statistic-based time series prediction models such as ARIMA \cite{williams2003modeling} and SVR \cite{castro2009online} are widely used for predicting traffic flows on a single road. Due to the limited learning capacity, these statistic-based approaches cannot effectively capture the complex spatio-temporal dependencies of the traffic data over a large road network. Recently, various deep learning based methods are broadly applied in traffic flow prediction and these models have achieved much better performance than traditional statistic-based shallow models, such as ST-ResNet \cite{zhang2017deep} and ConvLSTM \cite{xingjian2015convolutional}. SeqST-GAN \cite{wang2020seqst} was proposed to perform multi-step traffic flow prediction of a city in an adversarial learning way. Above mentioned works mainly applied CNN to capture the spatial correlation by treating the traffic data of an entire city as images, or combined CNN and RNN models to capture both the spatial and temporal correlations. 

However, CNN based models cannot well model the semantic spatial correlation of the traffic data and the global spatial structure of a city. To address this issue, some recent works try to use graph neural network to perform road network level traffic prediction. \cite{li2018diffusion} proposed the diffusion Convolutional Recurrent Neural Network (DCRNN) to model the traffic flow as a diffusion process on a directed road graph. A Spatio-Temporal Graph Convolutional Networks was proposed to tackle the time series prediction problem in traffic forecasting. \cite{zheng2020gman} proposed a graph multi-attention network (GMAN) to predict traffic conditions for time steps ahead at different locations on a road network graph. However, complex spatial dependencies which contains local and global spatial features are not well considered in these models.
\begin{figure*}
    \centering
    \includegraphics[scale=0.48]{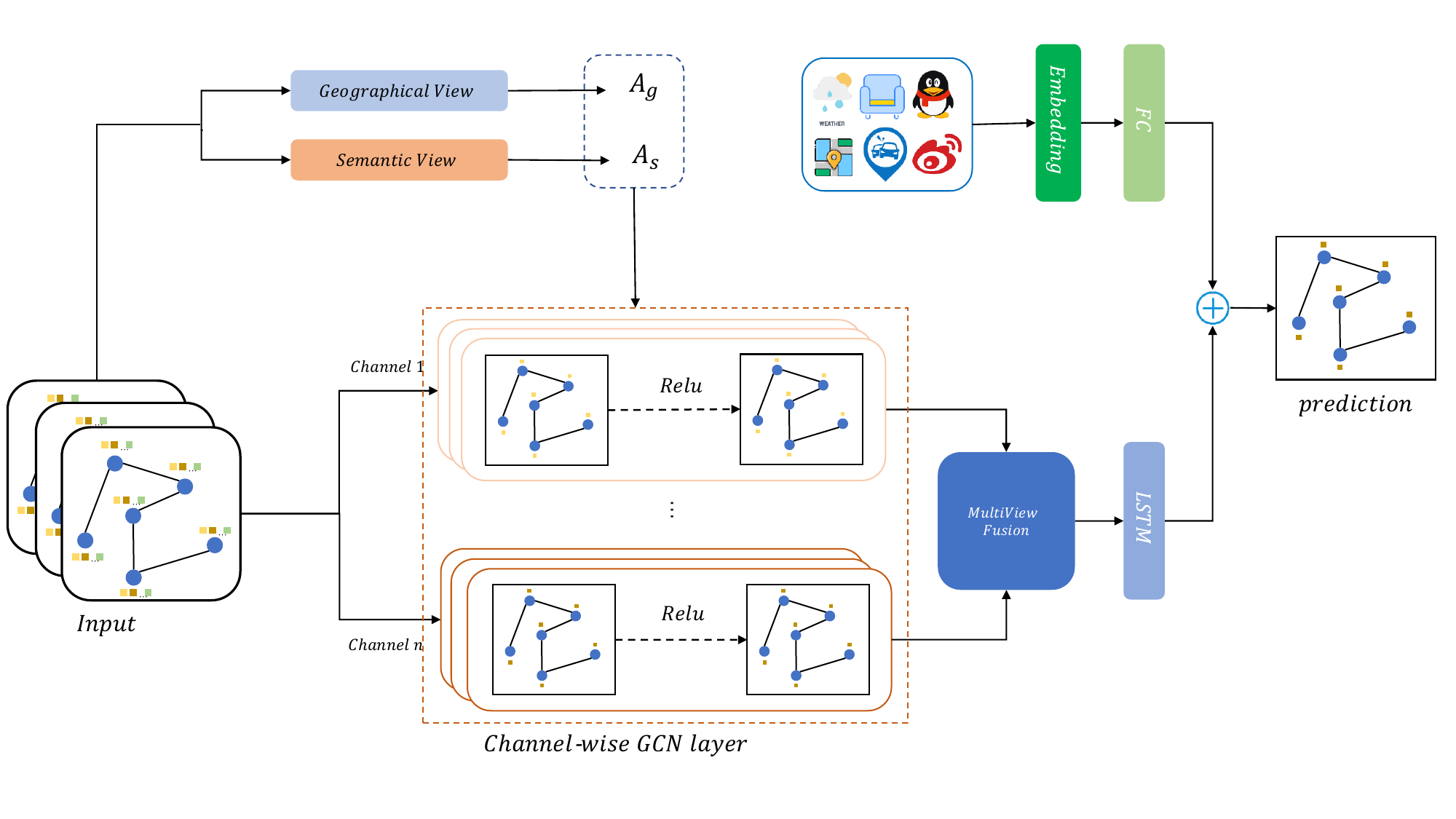}
    \caption{The framework of MVC-STNet.(Relu is a non-linear activation function)}
    \label{model}
\end{figure*}

\textbf{Graph Convolutional Network.} Graph convolutional network (GCN) was widely studied in recent years \cite{qu2019gmnn}. \cite{estrach2014spectral} first proposed the graph convolution operation in Fourier domain through the graph Laplacian. Then Chebyshev expansion of the graph Laplacian was employed to improve the inference efficiency \cite{defferrard2016convolutional}. \cite{kipf2017semi} simplified the convolution operation which only aggregated the node features from their neighbors. GAT \cite{velivckovic2018graph} introduced the attention mechanism to aggregate node features with the learned wights. GraphSAGE \cite{hamilton2017inductive} proposed a general, inductive framework that leveraged node feature information (e.g., text attributes) to efficiently generate node embedding.  However, these GCN based models treat the input features equally, but ignore the different impacts of node features on prediction tasks.
\section{Problem Formulation}
In this section, we will first give some definitions to help us state the studied problem. Then a formal problem formulation will be given.
\begin{definition}
\textbf{Traffic network $\mathcal{G}$} we denote the traffic network at time slot $t$ as $G^t = \{V^t, E^t, A^t\}$, where $|V^t|=N$ is the set of vertices, $N$ denotes the number of vertices that represent road segments in real world, $E^t$ denotes the set of edges, indicating the connectivity between the nodes, for example, if two segments in real world are connected, an edge exists between the corresponding nodes. Note that the traffic network can be either directed or undirected, and $A^t$ represents the adjacency matrix.

\end{definition}
\begin{definition}
\textbf{Traffic network feature matrix $\mathcal{X}$} we define the traffic network feature matrix at time slot t as $X^t \in \mathcal{R}^{N\times C}$, where $C$ is the dimension of the node features. The traffic network feature matrix represents the observations of $\mathcal{G}$ at the time step $t$. 
\end{definition}
\begin{problemdefinition}
Given the spatio-temporal traffic network and graph feature matrix $\{X^t, G^t|t=t_1, \dots, t_{T}\}$ over T time slots, and external context data matrix $E$ (e.g., weather, holiday, etc.), our goal is to to predict the traffic flow $Y^{t}$ in the next time slot.
\end{problemdefinition}

\section{Deep Multi-View Channel-wise Spatio-Temporal Network}
Figure \ref{model} shows the framework of the proposed MVC-STNet model. As shown in the figure, the model contains four major steps. First, we split the raw input traffic network data into $n$ parts with each part indicating a channel. The local and global spatial-temporal graphs are constructed from the geographical and semantic views, respectively. Next, several channel-wise GCN (CGCN) layers are stacked to learn the hidden feature of each channel which represents traffic flow, vehicle speed or road occupancy. From the picture \ref{relation}, we assume that different channels may have different impacts on task prediction. The stacked CGCN layers aim to capture the spatial dependencies of the data and map the channel-wise data into a high-dimensional embedding latent space. Meanwhile, it also can adaptively learn the hidden relations between input channels and prediction. Third, after the multi-view fusion, the learned spatial features are input into stacked LSTM layers to learn the temporal dependencies. Finally, we concatenate the learned spatio-temporal representations and the learned external features for prediction. Next, we will elaborate these steps in detail in the following subsections.

\subsection{Local and Global Spatial-Temporal Graph Construction}
We intend to build a model that can directly capture the influence of each node on its local neighbors and global neighbors. We use $A_{g} \in \mathcal{R}^{N\times N}$ and $A_{s} \in \mathcal{R}^{N\times N}$ to denote the local and global adjacency matrix of the spatial graph, respectively. Based on the first Law of Geography \cite{tobler1970computer}: "\textit{Near things are more related than distant things}", we first construct the
local spatial graph. If two nodes are connected with each other geographically, there is an edge between them. The corresponding value in the adjacency matrix is set to be the reciprocal of the distance between the two nodes. The geographical adjacency matrix can be formulated as follows:
\begin{equation}
A^{i,j}_g = \left\{
    \begin{array}{lr}
    \frac{1}{dis}, if\; v_i\; connects\; to\; v_j\\
    \;0, \ \, otherwise
    \end{array}
    \right.
\end{equation}
where $dis$ denotes the geographical distance between two nodes $v_i$ and $v_j$. 

However, the First Law of Geography may not fully reflect the spatial correlations of the traffic flows in urban areas that is known as the global spatial correlations or semantic spatial correlations. For example, a commercial area may have few traffic flows coming from a park, although they are geographically close to each other; while a residential district far away may have a large number of people flowing into the commercial area. To tackle this problem, we construct the adaptive global adjacency matrix which was used in AGCRN \cite{bai2020adaptive}. First, we randomly initialize a learnable node embedding: $E_A \in \mathcal{R}^{N \times d_e}$ for all nodes, where $d_e$ denotes the dimension of node embedding, and each row of $E_A$ represents the embedding of a specific node. Then we can infer the semantic spatial dependencies between each pair of nodes by multiplying $E_A$ and $E_A^T$ as follows:
\begin{equation}
    A_s = Softmax(ReLU(E_AE_A^T)
\end{equation}
where $Softmax$ is used to normalize the adaptive matrix. Note that the globalized adjacency matrix changes over time. This process is similar as constructing the graph based on nodes similarity.

\subsection{Multi-View Channel-wise Graph Convolutional Network}
The localized spatial graphs and the globalized spatial graphs are next input into the Multi-View Channel-wise Graph Convolutional Network for modeling the geographical and semantic spatial correlations. As the input features are treated equally by the traditional GCN \cite{kipf2017semi}, the different influences of various features on the prediction is ignored. To address this issue, we propose a channel-wise graph convolutional network (CGCN) to first learn the data representation of each channel separately, and then fuse them together. Moreover, we propose a multi-view fusion module for modeling both localized and globalized spatial correlations, which can be formulated as follows:
\begin{equation}
    \begin{split}
        &H_g^t  = CGCN(X^t, A_g)\\
        &H_s^t  = CGCN(X^t, A_s)\\
        &H_{mv}^t  = H_g^t + H_s^t
    \end{split}
    \label{MVCGCN}
\end{equation}
where $H_g^t$, $H_s^t$ and $H_{mv}^t$ are localized spatial hidden features, globalized spatial hidden features and multi-view fused spatial features respectively, $CGCN(\cdot)$ denotes the proposed channel-wise graph convolution network.
\begin{figure}[htb]
    \flushleft
    \includegraphics[scale=0.34]{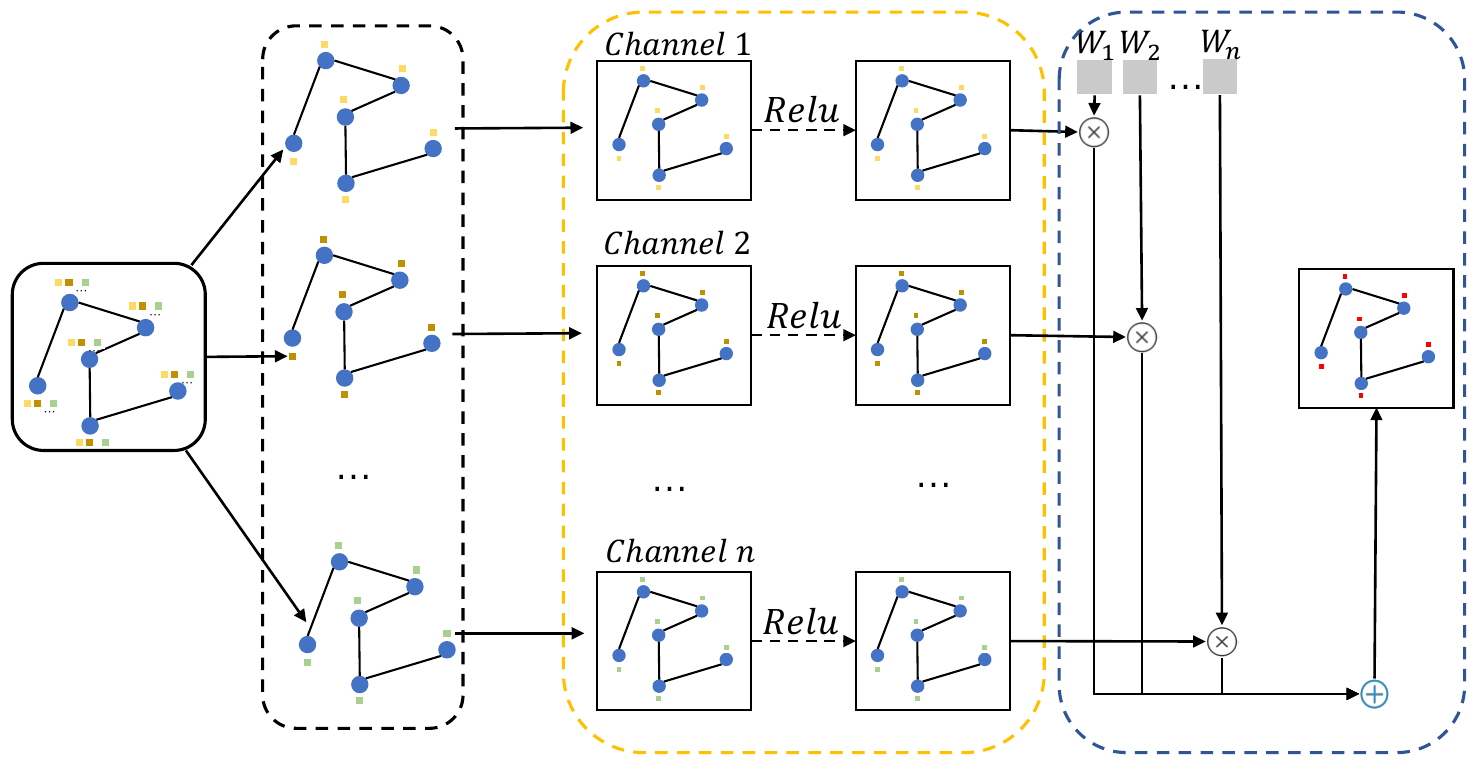}
    \caption{The illustration of channel-wise graph convolutional network.(Relu is a non-linear activation function)}
    \label{cgcn}
\end{figure}
\subsubsection{Channel-wise Graph Convolutional Network}
Recently, generalizing convolutional networks to graph data have attracted considerable research interest. In this paper, we consider to use spectral convolutions \cite{kipf2017semi} on the constructed spatial graphs, which can be formulated as follows:
\begin{equation}
   f(X^t, A)=\sigma(D^{-\frac{1}{2}}\widetilde{A} D^{-\frac{1}{2}}X^{t}W^{t})
\end{equation}
where $f(\cdot)$ represents the GCN operation, $\widetilde{A} = A + I_N$ is the adjacency matrix of $\mathcal{G}$ with added self-connections, $\widetilde{D}_{ii}=\sum_j \widetilde{A}_{ij}$ is the degree matrix , $W^t$ represents the learnable weight matrix, and $\sigma(\cdot)$ is the activation function.

However, the above GCN operation cannot model the different influences of various input features on the final prediction. To address this issue, we propose a channel-wise graph convolutional network (CGCN) to separately model relations between each input features and output. Figure \ref{cgcn} shows the framework of the proposed CGCN. We first split the input graph features $X^t \in \mathcal{R}^{N\times C}$ by channels to get the channel-wise spatial graph features $X^t_i \in \mathcal{R}^{N\times 1}$. Then we input the $X^t_i$ into the stacked GCN layer which is shared by all channel-wise graph features. The learned $i$-$th$ localized spatial hidden features can be formulated as follows: 
\begin{equation}
    H_{i,g,l+1}^t = \sigma(D^{-\frac{1}{2}}\widetilde{A_g} D^{-\frac{1}{2}}H^{t}_{i,g,l}W^{t}_l)
\end{equation}
where $H_{g,0}^t$ is equal to $X^t_i$, $W_l^t$ is the a trainable matrix of filter parameters in the $l$-$th$ graph convolutional layer. We use $CGCN_i(X_i^t, A_g)$ to denote the localized GCN operation. Similarly, the globalized graph convolutional operation for $i$-$th$ channel can be defined as:
\begin{equation}
    H_{i,s,l+1}^t = \sigma(D^{-\frac{1}{2}}A_s D^{-\frac{1}{2}}H^{t}_{i,s,l}W^{t}_l)
\end{equation}
We use $CGCN_i(X_i^t, A_s)$ to represent the globalized GCN operation.\\
As shown in Figure \ref{relation}, with the change of traffic flow, the speed and occupancy show different trends. Inspired by this observation, We propose a novel channel-fusion method to fuse the latent representations of all channel features. This method aims to automatically learn the relations between outputs and each input features. We employ the parametric-matrix-based \cite{zhang2017deep} fusion method to fuse the $n$ outputs of stacked CGCN layers as follows:
\begin{equation}
    \begin{split}
        H_g^t = W_1 \odot H_{1,g}^t + W_2 \odot H_{2,g}^t + \dots + W_n \odot H_{n,g}^t\\
        H_s^t = W_1 \odot H_{1,s}^t + W_2 \odot H_{2,s}^t + \dots + W_n \odot H_{n,s}^t
    \end{split}
\end{equation}
where $W_i(i\in \{1,\dots,n\})$ are the learnable parameters that adjust the degrees affected by $n$ channels, $H_{\alpha \beta}^t(\alpha \in \{1, \dots, n\}, \beta \in \{g,s\})$ are the learned spatial representations of $n$ channels' feature. 

\subsection{Learning Temporal Dependencies with LSTM}
Besides the spatial correlations, traffic flow forecasting also involves complex temporal correlations. We input the extracted spatial features into LSTM layers as follows for temporal feature learning.
\begin{equation}
    \begin{split}
        &f^t = \sigma(W_f[h^{t-1},H^t_{mv}]+b_f),\\
  &i^t = \sigma(W_i[h^{t-1},H^t_{mv}]+b_i),\\
  &c^t = f_t\odot c^{t-1}+i_t\odot tanh(W_c[h^{t-1},H^t_{mv}] + b_c),\\
  &o^t = \sigma(W_o[h^{t-1},H^t_{mv}]+b_o),\\
  &h^t = o^t\odot tanh(c^t). 
    \end{split}
    \label{lstm}
\end{equation}
where $f^t, i^t, c^t, o^t, h^t$ are forget gate, input gate, memory cell, output gate and hidden state, respectively.

\subsection{Integrating the External Features}
External context features may significantly affect the traffic flows. For example, the patterns of traffic on weekdays and weekends can be quite different, while rainstorms may dramatically decrease the traffic flows. By considering the external context features including weather conditions and holidays, we design an external feature extractor which is defined as :
\begin{equation}
    e^t = FC(Embedding(E^t))
    \label{external}
\end{equation}
where $e^t$ is the learned external feature representation, $FC(\cdot)$ represents the multi-layer perception model. Finally, we concatenate the learned spatial-temporal graph representations and external feature representation for traffic flow prediction.
\begin{equation}
    \hat{Y}^t = FC(Concat(h^t, e^t))
    \label{predict}
\end{equation}
where $Concat(\cdot)$ denotes the concatenation operation.
\subsection{Overall Objective Function}
The final loss function of MVC-STNet is as follows:
\begin{equation}
    Loss = \frac{1}{L}\sum_{t=1}^{L}||\hat{Y}^t - Y^t||^2
\end{equation}
where $L$ is the training sample size, $\hat{Y}^t$ is the prediction and $Y^t$ is the ground truth. The pseudo-code of the algorithm is shown in Algorithm \ref{algorithm}.

\begin{algorithm}[!t]
	\caption{\small Deep Multi-View Channel-wise Spatio-Temporal Network}
	\label{algorithm}
	\begin{algorithmic}[1]
		\Require
		$\mathcal{G}$: Spatial-temporal traffic network $X$: Spatial-temporal graph signal matrix
		\Ensure Parameter set
		$\Theta$
		\State Initialize parameters $\Theta$
		
		\While {$not$ $converge$}
		\State 0 $\leftarrow$ $t$, $t$ is the $t$-$th$ time slot
		\While{$t<T$}
		\State Sample {$X^T\in X, \mathcal{G}^t\in \mathcal{G}$} 
		\State $H_{mv}^t \leftarrow$ Learning spatial correlations with multi-view channel-wise GCN through Eq. \ref{MVCGCN}
		\State $h_t \leftarrow$ Learning temporal correlations with LSTM by Eq. \ref{lstm}
		\State $e^t \leftarrow$ External feature learning by Eq.\ref{external}
		\State $\hat{Y}^t \leftarrow$ Predict traffic flows with Eq.\ref{predict}
		\State update $\Theta$ based on Objective function
		\State $t \leftarrow t+1$
		\EndWhile
		
		\State \textbf{return} $\Theta$
		\EndWhile
	\end{algorithmic}
\end{algorithm}
\begin{table}[!tbp]
    \centering
    \caption{Dataset Description}
    \scalebox{0.78}{
    \begin{tabular}{ccc}
    \hline
         Dataset& PeMS04 & PeMS08 \\
         \hline
         Area& San Francisco Bay & San Bernaridino\\ 
         Detectors & 3848 & 1979\\
         Time span &01/01/2018$\sim$28/02/2018 &01/07/2016$\sim$31/08/2016\\
         Time interval & 5 mins & 5 mins\\
         Nodes &307 & 170\\
         \# of time intervals & 16992& 17847\\
         \hline
         \multicolumn{3}{c}{\textbf{External Features}}\\
         Days & \multicolumn{2}{c}{Weekday, weekend, holiday, etc}\\
         Weather conditions & \multicolumn{2}{c}{Temperature, rain, snow, etc.}\\
    \hline
    \end{tabular}}
    \label{dataset}
\end{table}
\section{Experiment}
\begin{table*}
	\small
	\centering
	\caption{RMSE and MAE comparison among different methods}
	\setlength{\tabcolsep}{9mm}{
	\begin{tabular}{c|c|c|c|c}
		\hline \multirow{1}{*} { Model } & \multicolumn{2}{|c|} { PeMS04 } & \multicolumn{2}{|c} { PeMS08 } \\
		\cline { 2 - 5 } & \multicolumn{1}{|c|} { RMSE } & \multicolumn{1}{|c|} { MAE } & \multicolumn{1}{|c} { RMSE } & \multicolumn{1}{|c} { MAE } \\
		\hline HA & 47.80 & 32.73& 43.32 & 30.09  \\
		ARIMA & 55.18 & 36.84 & 48.88 & 34.27  \\
		FC-LSTM & 43.18 & 29.52 & 39.94 & 26.80  \\
		STGCN & 42.37 & 30.69 & 37.83 & 27.66  \\
		DCRNN & 39.86& 26.78& 33.62& 25.44\\
		T-GCN & 39.45 & 27.69 & 34.43 & 24.92 \\
		MVC-STNet & \textbf{37.61} & \textbf{23.83} & \textbf{32.88} & \textbf{21.40} \\
		\hline
	\end{tabular}}
	\label{baselines}
\end{table*}
\subsection{Dataset and Experiment Setup}
\subsubsection{Dataset} We use two datasets that are widely used in graph traffic flow prediction for evaluation: PeMS04, and PeMS08. The details of the datasets are introduced as follows.

 \textbf{\textit{PeMS04}} This traffic dataset is collected in San Francisco Bay Area. It contains the traffic data collected by 3848 detectors on 29 roads. The time span of this dataset is from January to February in 2018. We use the first 54 days data for training and validating, and the remaining data are for testing.
 
 \textbf{\textit{PeMS08}} This dataset is collected in San Bernaridino from July to August in 2016. It contains the traffic data collected by 1979 detectors on 8 roads. We use the first 55 days for training and validating,and the remaining 7 days data for testing.
 
We also use some external features including weather, holiday and weekends. Whether the day is weekday, weekend or holiday is also considered as the people mobility patterns on holidays and regular days are quite different. The descriptions on the two datasets and external features are shown in Table \ref{dataset}.
\subsubsection{Baselines} We compare the proposed MVC-STNet with the following baseline methods.
\begin{itemize}
    \item \textbf{HA} Historical Average(HA) uses the average historical data as the prediction of the future.
    \item \textbf{ARIMA} Auto-Regressive Integrated Moving Average (ARIMA) \cite{williams2003modeling} is a classic statistics-based method for time series prediction.
    \item \textbf{FC-LSTM} FC-LSTM \cite{sutskever2014sequence} uses RNN with fully connected LSTM hidden units to capture the non-linear temporal dependencies for traffic prediction.
    \item \textbf{STGCN} STGCN \cite{yu2018spatio} applies ChebNet-GCN and 1D convolution to extract spatial and temporal dependencies for traffic prediction.
    \item \textbf{DCRNN} DCRNN \cite{li2018diffusion} uses graph convolution networks to capture the spatial correlations and the encoder-decoder architecture with scheduled sampling to capture the temporal correlations for traffic prediction.
    \item \textbf{T-GCN} T-GCN \cite{zhao2019t} combines the graph convolutional network (GCN) and the gated recurrent unit (GRU) for traffic prediction
\end{itemize}
To further evaluate the effectiveness of different components in our model, we also compare the full version MVC-STNet with the following variants:
\begin{itemize}
    \item \textbf{GCN-STNet} This model removes the Channel-wise  Convolutional network and the multi-view fusion module. It only considers the local spatial correlations. By comparing with it, we test whether the proposed $CGCN(\cdot)$ and multi-view fusion is useful for improving the prediction performance.
    \item \textbf{CGCN-STNet} This model does not consider the features of the globalized spatial-temporal graph. Through comparing with this model, we test whether integrating the global spatial graph is helpful to capture the complex spatial features. 
    \item \textbf{MV-STNet} This model drops the CGCN layer while only use the tradiction GCN layer \cite{kipf2017semi} to capture the spatial correlations. Through comparing with it, we test whether the proposed CGCN method can model the divergence between channels.
\end{itemize}

\subsubsection{Evaluation metric} We adopt mean absolute error (MAE) and root mean square error (RMSE) defined as follows as the evaluation metrics.
\begin{equation}
\small
    MAE = \frac{1}{n}\sum_{t=1}^{n}|\hat{Y_t}-Y_t|,
    RMSE = \sqrt{\frac{1}{n}\sum_{t=1}^n||\hat{Y_t}-Y_t||^2}
\end{equation}
where $\hat{Y}^t$ is the prediction, and $Y^t$ is the ground truth.
\begin{figure}
    \centering
    \includegraphics[scale=0.42]{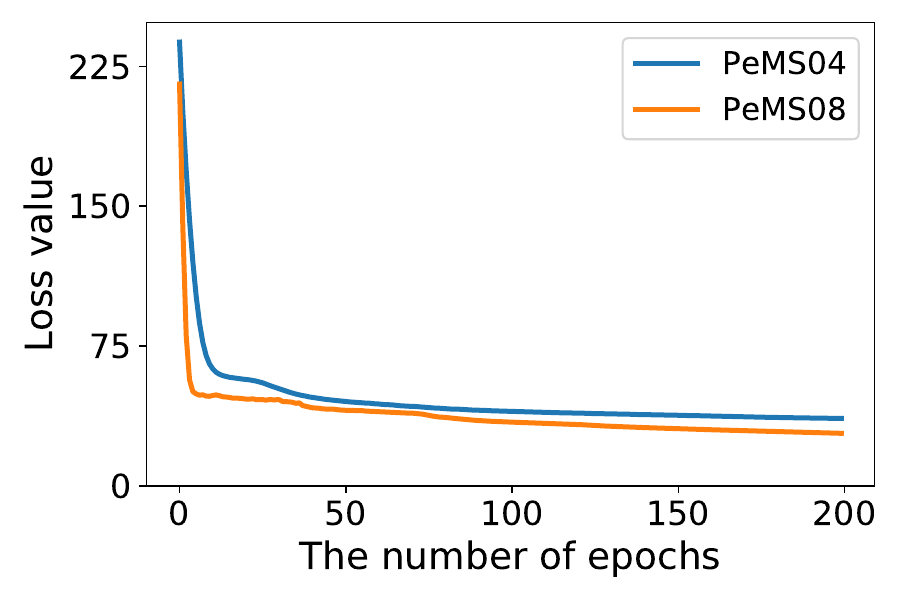}
    \caption{Loss curves of MVC-STNet on the two datasets}
    \label{loss}
\end{figure}
\subsubsection{Implement details} We implement our model with Pytorch framework on NVIDIA RTX3090 GPU. The model parameters are set as follows. The input data size for PeMS04 dataset is $3\times 307 \times3$ where the first 3 is the previous time slot length used for prediction, 307 is the number of the nodes, the last 3 is the number of channels that represent flow, speed and occupy, for PeMS08 dataset is $3 \times 170 \times 3$, where 170 represents the number of the nodes. The CGCN model contains 3 layers whose hidden feature dimensions are $16, 32,$ and $64$. The final LSTM contains 2 layers, whose hidden dimension are all 256. The output of PeMS04 is $1\times 307\times 1$, and the output of PeMS08 is $1 \times 170 \times 1$. The baseline methods are implemented based on the original papers or we use the publicly available code. The parameters of baseline methods are set based on the original papers. Note that we normalize the traffic data into $[0, 1]$ to facility the feature learning.

\subsubsection{Convergence of the algorithm} Figure \ref{loss} shows the training loss curves of the algorithm on the two datasets. One can see that MVC-STNet converges after about 50 epochs on the two datasets, and then becomes stable. It shows the proposed model can quickly converge. In the following experiments, we run MVC-STNet with 50 epochs on both datasets.

\subsection{Comparison with Baselines}
Table \ref{baselines} shows the performance comparison among different methods over the two datasets. The best results are highlighted with bold font. It shows that the proposed MVC-STNet achieves the best performance over both datasets. Traditional statistics-based methods ARIMA and HA achieve the worse performance among all the methods. It is not surprising because ARIMA and HA only use the time series data of each node, but ignore the spatial dependency. On PeMS04 dataset, compared with the best results achieved by baselines, MVC-STNet reduces RMSE of the traffic flow prediction from 39.45 achieved by T-GCN to 37.61, and MAE from 26.78 achieved by DCRNN to 23.83. MVC-STNet reduces the RMSE from 33.62 to 32.88, and MAE from 24.92 to 21.40. The decrease trends of RMSE and MAE over PeMS08 dataset are smaller than that over PeMS04. This is because the number of nodes in the traffic network of PeMS08 is smaller than that of PeMS04, which means that the spatial correlations is more complex in PeMS04. The results in Table \ref{baselines} show that the proposed MVC-STNet is superior to existing state-of-the-art spatio-temporal learning approaches.

\subsection{Comparison with Variant Models}
To study whether the components in MVC-STNet are all helpful to the prediction task, we compare with its variants GCN-STNet, CGCN-STNet, and MV-STNet. The result is shown in Table \ref{variant}. One can that CGCN and multi-view fusion module are all useful to the model as removing any one of them will increase the prediction error. On PeMS04 dataset, considering the divergence between channels seems more important, while the multi-view fusion operation is more important on PeMS08 dataset. Combining these components together achieves the lowest RMSE and MAE, demonstrating that all of them are useful to the studied problem.
\begin{table}[]
    \centering
    \small
    \setlength{\tabcolsep}{4mm}{
    \begin{tabular}{cccc}
         \hline
         Dataset& Methods & RMSE & MAE \\
         \hline
         \multirow{4}{*}{PeMS04} & GCN-STNet &40.77&26.83\\
         & CGCN-STNet &38.47&24.83\\
         & MV-STNet &38.91&25.12\\
         & MVC-STNet &37.61&23.83\\
         \hline
         \multirow{4}{*}{PeMS08} & GCN-STNet & 37.33& 25.63\\
         & CGCN-STNet &34.48&22.93\\
         & MV-STNet &33.90&22.44\\
         & MVC-STNet &32.88&21.40\\
         \hline
    \end{tabular}}
    \caption{RMSE and MAE comparison with variant methods}
    \label{variant}
\end{table}
\subsection{Case Study}
\begin{figure}
    \flushleft
    \includegraphics[scale=0.21]{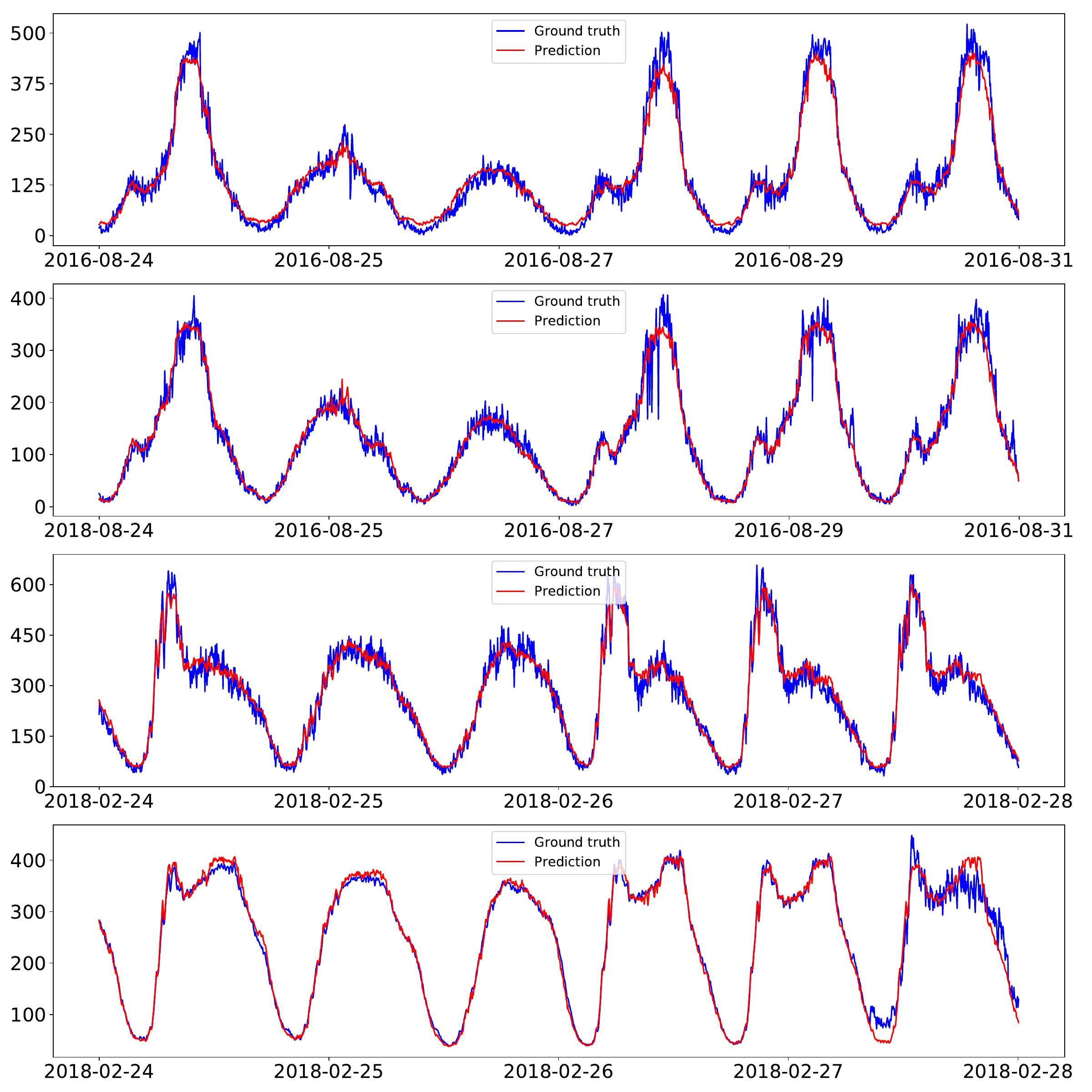}
    \caption{Prediction vs ground truth on two datasets(top to down: node 7 and node 15 in PeMS04, node 20 and node 29 in PeMS08)}
    \label{casestudy}
\end{figure}

To further intuitively illustrate how accurately our model can predict the traffic flows, we visualize the predicted traffic flows and the ground truth in Figure \ref{casestudy}. Due to space limitation, we only show a case study on the selected four nodes of the traffic network on the two datasets. From top to down, the upper two figures show the predicted traffic flows and the ground truth on nodes 7 and 15 of the PeMS04 dataset, and the lower two figures show the prediction and the ground truth on the nodes 20 and 29 of the PeMS08 dataset. One can see that the red curves of prediction can accurately trace the blue curves of the ground truth. The figure also shows that the two datasets present obvious periodical change characteristics, which is consistent with the traffic mobility patterns in cities. Our model can perfectly capture the periodicity of the data due to the usage of LSTM. Furthermore, the results show that our proposed MVC-STNet model can capture the spatial and temporal dependencies effectively. 
\section{Conclusion}
In this paper, we propose a novel deep multi-view channel-wise spatio-temporal network named MVC-STNet. The novelty of the model lies in the proposed channel-wise graph convolutional network to learn the divergence between channels. The local and global spatio-temporal graph are constructed to capture both the geographical and semantic spatial correlations. Extensive evaluations on two real datasets show that the proposed model can mutually enhance the performance of both tasks, and also outperforms state-of-the-art graph based deep learning methods for traffic flow prediction. In the future, we will further study how to learn more temporal dependencies to capture sudden changes of traffic flows. In addition, it would be also interesting to further study how the proposed MVC-STNet framework can be applied to other prediction tasks such as traffic congestion prediction and traffic accident prediction.
\section{Acknowledgement}
This work is supported by CCF-Tencent Open Research Fund, and the Fundamental Research Funds for the Central Universities (No.: NZ2020014).
\bibliographystyle{aaai}
\bibliography{refer}
\end{document}